\relax
\documentclass[letterpaper]{article} 
\usepackage{aaai21}  
\usepackage{times}  
\usepackage{helvet} 
\usepackage{courier}  
\usepackage[hyphens]{url}  
\usepackage{graphicx} 
\usepackage{algorithm}
\usepackage{algorithmic}
\usepackage{amssymb}
\usepackage{amsmath}
\usepackage[backref]{hyperref} 

\usepackage{epsfig}
\usepackage{graphicx}
\usepackage{multirow}
\usepackage{tabularx}
\usepackage{booktabs}
\usepackage{url}
\usepackage{makecell}

\usepackage{float}
\usepackage{stfloats}
\usepackage{subfigure}
\usepackage{graphicx}
\usepackage{listings}
\usepackage[hyphens]{url}

\usepackage{natbib}  
\usepackage{caption} 
\title{PP-StructureV2: A Stronger Document Analysis System}
\author {
    Chenxia Li, Ruoyu Guo, Jun Zhou,  Mengtao An, \\
   Yuning Du, Lingfeng Zhu, Yi Liu, Xiaoguang Hu, Dianhai Yu \\
}
\affiliations {
    Baidu Inc. \\
    \{lichenxia, zhulingfeng\}@baidu.com
}

\begin{document}

\maketitle

\begin{abstract}

A large amount of document data exists in unstructured form such as raw images without any text information. Designing a practical document image analysis system is a meaningful but challenging task. In previous work, we proposed an intelligent document analysis system PP-Structure. In order to further upgrade the function and performance of PP-Structure, we propose PP-StructureV2 in this work, which contains two subsystems: Layout Information Extraction and Key Information Extraction. Firstly, we integrate Image Direction Correction module and Layout Restoration module to enhance the functionality of the system. Secondly, 8 practical strategies are utilized in PP-StructureV2 for better performance. For Layout Analysis model, we introduce ultra light-weight detector PP-PicoDet and knowledge distillation algorithm FGD for model lightweighting, which increased the inference speed by 11 times with comparable mAP. For Table Recognition model, we utilize PP-LCNet, CSP-PAN and SLAHead to optimize the backbone module, feature fusion module and decoding module, respectively, which improved the table structure accuracy by 6\% with comparable inference speed. For Key Information Extraction model, we introduce VI-LayoutXLM which is a visual-feature independent LayoutXLM architecture, TB-YX sorting algorithm and U-DML knowledge distillation algorithm, which brought 2.8\% and 9.1\% improvement respectively on the Hmean of Semantic Entity Recognition and Relation Extraction tasks. All the above mentioned models and code are open-sourced in the GitHub repository PaddleOCR \footnote{\url{https://github.com/PaddlePaddle/PaddleOCR/tree/release/2.6/ppstructure}}.

\end{abstract}

\section{Introduction}

Document intelligence is a booming research topic and practical industrial demand in recent years. It mainly refers to the process of understanding, classification, extraction and information induction through artificial intelligence technology for the text and rich typography contained in web pages, digital documents or scanned documents. Due to the diversity of layouts and formats, low-quality scanned document images, and the complexity of template structures, document intelligence is a very challenging task and has received extensive attention in related fields. Layout Analysis, Table Recognition, and Key Information Extraction are three representative tasks in intelligent document analysis.

Document Layout Analysis can be regarded as an object detection task for document images in essence. The basic units such as titles, paragraphs, tables, and illustrations in the document are the objects needed to be detected and recognized. Layout-parser\cite{layoutparser} is a unified toolkit for Deep Learning Based Document Image Analysis. VSR\cite{vsr} is proposed for layout analysis, which comes to state-of-the-art on PubLayNet dataset\cite{publaynet}. In PP-Structure, we use PP-YOLOv2\cite{ppyolov2} to complete the layout analysis task, which is real-time on GPU devices. However, currently proposed models are not CPU-friendly and thus not conducive to deployment on CPUs or mobile devices.

Table Recognition is used to convert table images into editable Excel format files. The diversity of tables in document images, such as various rowspans and colspans and different text types, makes table recognition a hard task in document understanding. There are many table recognition methods, such as traditional algorithms based on heuristic rules and recently developed methods based on deep learning. Among them, the end-to-end method has received extensive attention due to the simplicity of the pipeline, which represent the table in HTML format and adopt Seq2Seq\cite{seq2seq} to predict the table structure, such as TableRec-RARE\cite{TableRec-RARE} in PP-Structure powered by PaddlePaddle\cite{paddlepaddle}. In TableMaster\cite{TableMaster}, transformer is used as the decoder, which achieves high accuracy, but brings huge computation cost.

Key Information Extraction (KIE) refers to extracting the specific information that users pay attention to. Semantic Entity Recognition (SER) and Relation Extraction (RE) are two main subtasks for KIE. LayoutLM\cite{layoutlm} is firstly proposed to jointly model interactions between text and layout information across scanned document images, which is beneficial to the downstream KIE process. LayoutLMv2\cite{layoutlmv2} integrates the image information in the pre-training stage by taking advantage of the transformer architecture to learn the cross-modality interaction between visual and textual information. LayoutXLM\cite{layoutxlm} is a multilingual extension of LayoutLMv2\cite{layoutlmv2} model. XY-LayoutLM \cite{xylayoutlm} proposed Augmented XY-CUT algorithm to sort the textlines in human reading order based on the observation that reading order is vital for KIE. However, these multi-modal approaches do not pay much attention to inference time.

\begin{figure*}[ht]
\centering
\subfigure{
\includegraphics[width=0.9\textwidth]{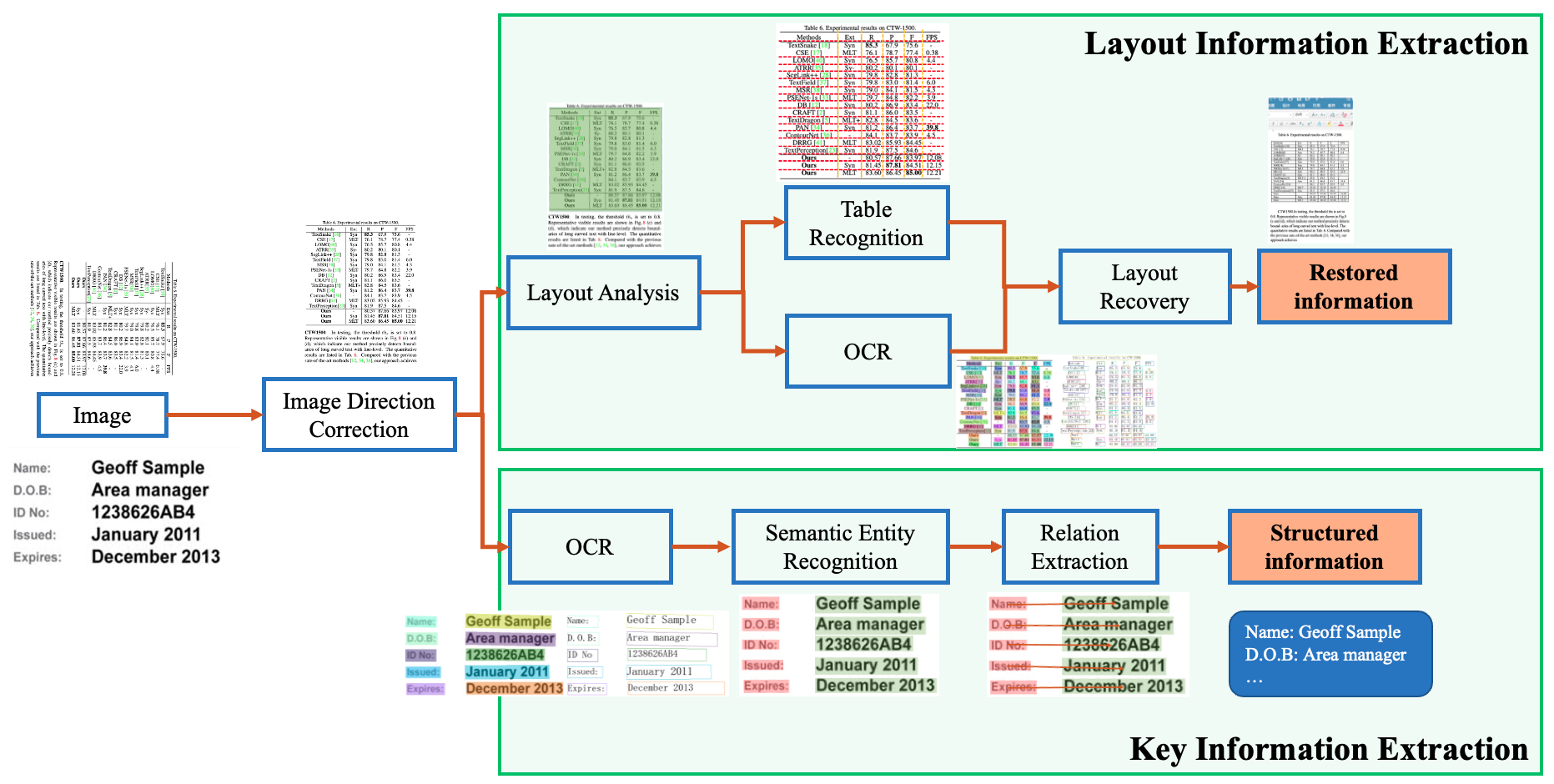}
}
\caption{Framework of the proposed PP-StructureV2. It contains two subsystems: layout information extraction and key information extraction.}
\label{ppstructurev2_framework}
\end{figure*}

PP-Structure is our first attempt for an intelligent document analysis system, which supports basic functions such as Layout Analysis and Table Recognition, but lacks consideration of efficiency, and there is still much room for performance improvement. In this work, we propose PP-StructureV2, a more robust and comprehensive document analysis system. Figure \ref{ppstructurev2_framework} shows the PP-StructureV2 framework. Firstly, the input document image direction is corrected by the Image Direction Correction module. For the Layout Information Extraction subsystem, as shown in the upper branch, the corrected image is firstly divided into different areas such as text, table and image through the layout analysis module, and then these areas are recognized respectively. For example, the table area is sent to the table recognition module for structural recognition, and the text area is sent to the OCR engine for text recognition. Finally, the layout recovery module is used to restore the image to an editable Word file consistent with the original image layout. For the Key Information Extraction subsystem, as shown in the lower branch, OCR engine is used to extract the text content, then the Semantic Entity Recognition module and Relation Extraction module are used to obtain the entities and their relationship in the image, respectively, so as to extract the required key information.

The contributions of this paper are summarized as follows:

\begin{itemize}
    \item We upgrade the intelligent document analysis system PP-Structure and proposed PP-StructureV2 with better performance.
    \item We newly introduce two modules in PP-StructureV2: Image Direction Correction and Layout Recovery, which support processing rotated images and restore images to editable Word files based on analysis results.
    \item We optimize Layout Analysis, Table Recognition and Key Information Extraction models, significantly surpassing the previous version in terms of speed or accuracy.
\end{itemize}

The rest of the paper is organized as follows. In section 2, we present the details of the newly proposed improvement strategies. Experimental results are discussed in section 3 and conclusions are conducted in section 4.

\section{Improvement Strategies}
 
\subsection{Image Direction Correction Module}

\begin{figure}[]
\centering
\subfigure{
\centering
\includegraphics[width=\columnwidth]{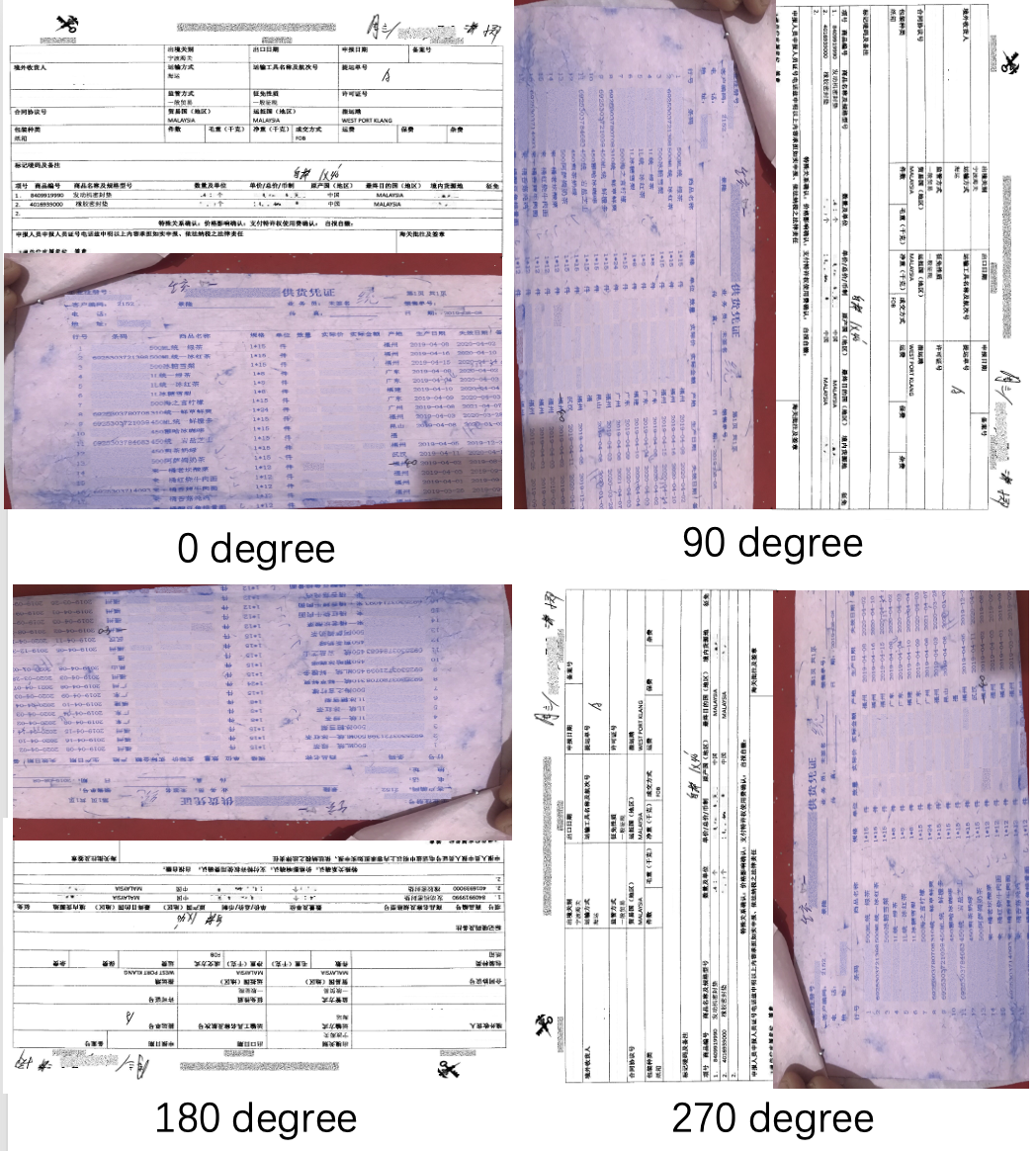}
}
\caption{Some images in the text image direction dataset.}
\label{text_image_orientation_data_demo}
\end{figure}

Since the training set is generally dominated by 0-degree images, the information extraction effect of rotated images is often compromised. In PP-StructureV2, the input image direction is firstly corrected by the PULC text image direction model\cite{PULC_text_image_orientation} provided by PaddleClas \footnote{https://github.com/PaddlePaddle/PaddleClas}. Some demo images in the dataset are shown in Figure \ref{text_image_orientation_data_demo}. Different from the text line direction classifier, the text image direction classifier performs direction classification for the entire image. The text image direction classification model achieves 99\% accuracy on the validation set with 463 FPS on CPU device.

\subsection{Layout Analysis}

Layout Analysis refers to dividing document images into predefined areas such as text, title, table, and figure. In PP-Structure, we adopted the object detection algorithm PP-YOLOv2\cite{ppyolov2} as the layout detector. In PP-StructureV2, we use a more lightweight detector PP-PicoDet\cite{picodet}, which achieves superior performance on mobile devices. In addition, we adjust the image scale for the layout analysis scene, and use a knowledge distillation algorithm named FGD\cite{fgd} to further improve the model accuracy.

\subsubsection{PP-PicoDet: A better real-time object detector on mobile devices}

PaddleDetection \footnote{https://github.com/PaddlePaddle/PaddleDetection} proposed a new family of real-time object detectors, named PP-PicoDet, which achieves superior performance on mobile devices. PP-PicoDet adopts the CSP structure to constructure CSP-PAN as the neck, SimOTA as label assignment strategy, PP-LCNet as the backbone, and an improved detection One-shot Neural Architecture Search(NAS) is proposed to find the optimal architecture automatically for object detection. We replace PP-YOLOv2 adopted by PP-Structure with PP-PicoDet, and 
adjust the input scale from 640*640 to 800*608, which is more suitable for document images. With 1.0x configuration, the accuracy is comparable to PP-YOLOv2, and the CPU inference speed is 11 times faster.

\subsubsection{FGD: Focal and Global Knowledge Distillation}

FGD\cite{fgd}, a knowledge distillation algorithm for object detection, takes into account local and global feature maps, combining focal distillation and global distillation. Focal distillation separates the foreground and background of the image, forcing the student to focus on the teacher’s critical pixels and channels. Global distillation rebuilds the relation between different pixels and transfers it from teachers to students, compensating for missing global information in focal distillation. Based on the FGD distillation strategy, the student model (LCNet1.0x based PP-PicoDet) gets 0.5\% mAP improvement with the knowledge from the teacher model (LCNet2.5x based PP-PicoDet). Finally the student model is only 0.2\% lower than the teacher model on mAP, but 100\% faster.

\subsection{Table Recognition}

\begin{figure*}[t]
\centering
\includegraphics[width=15cm]{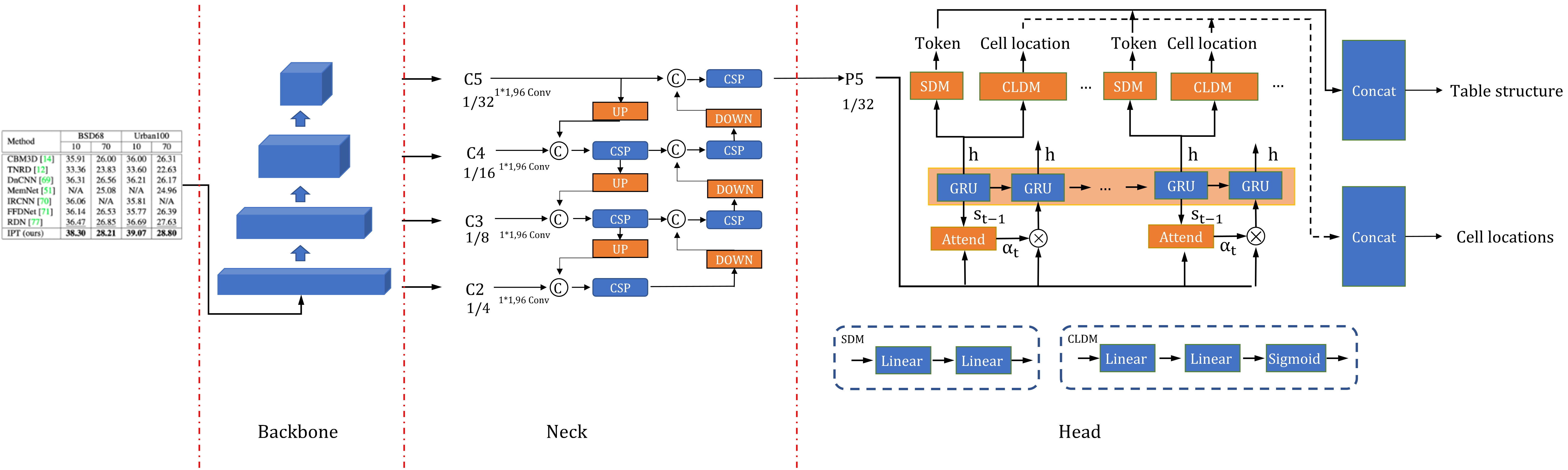}
\caption{Architecture of our proposed SLANet, where C represent concat operation.}
\label{slanet_framework}
\end{figure*}

In recent years, many Table Recognition algorithms based on deep learning have been proposed. In PP-Structure, we proposed an end-to-end Table Recognition algorithm TableRec-RARE\cite{TableRec-RARE}, based on the text recognition algorithm RARE\cite{rare}. The model output is an HTML representation of a table structure, which can be easily converted into Excel files. In PP-StructureV2, we propose an efficient Table Recognition algorithm named SLANet (\textbf{S}tructure \textbf{L}ocation \textbf{A}lignment Network). Compared with TableRec-RARE, SLANet has been upgraded in terms of model structure and loss. Figure \ref{slanet_framework} shows the network structure of SLANet.

\subsubsection{PP-LCNet: CPU-friendly Lightweight Backbone}

PP-LCNet\cite{PP-LCNet} is a lightweight CPU network based on the MKLDNN acceleration strategy, which achieves better performance on multiple tasks than lightweight models such as ShuffleNetV2\cite{shufflenet}, MobileNetV3\cite{mobilenetv3}, and GhostNet\cite{ghostnet}. Additionally, pre-trained weights trained by SSLD\cite{SSLD} on ImageNet are used for Table Recognition model training process for higher accuracy.

\subsubsection{CSP-PAN: Lightweight Multi-level Feature Fusion Module}

Fusion of the features extracted by the backbone network can effectively alleviate problems brought by scale changes in complex scenes. In the early days, the FPN\cite{FPN} module was proposed and used for feature fusion, but its feature fusion process was one-way (from high-level to low-level), which was not sufficient. CSP-PAN\cite{PP-PicoDet} is improved based on PAN. While ensuring more sufficient feature fusion, strategies such as CSP block and depthwise separable convolution are used to reduce the computational cost. In SLANet, we reduce the output channels of CSP-PAN from 128 to 96 in order to reduce the model size.

\subsubsection{SLAHead: Structure and Location Alignment Module}

In the TableRec-RARE head, output of each step is concatenated and fed into SDM (Structure Decode Module) and CLDM (Cell Location Decode Module) to generate all cell tokens and coordinates, which ignores the one-to-one correspondence between cell token and coordinates. Therefore, we propose the SLAHead to align cell token and coordinates. In SLAHead, output of each step is fed into SDM and CLDM to get the token and coordinates of the current step, the token and coordinates of all steps are concatenated to get the HTML table representation and coordinates of all cells. 

\subsubsection{Merge Token}

In TableRec-RARE, we use two separate tokens \textless td\textgreater \ and \textless/td\textgreater to represent a non-cross-row-column cell, which limits the network's ability to handle tables with a large number of cells. Inspired by TableMaster\cite{TableMaster}, we regard \textless td\textgreater \ and \textless/td\textgreater as one token - \textless td\textgreater\textless/td\textgreater in SLANet.

\subsection{Layout Recovery}

\begin{figure}[H]
\centering
\centering
\includegraphics[width=\columnwidth]{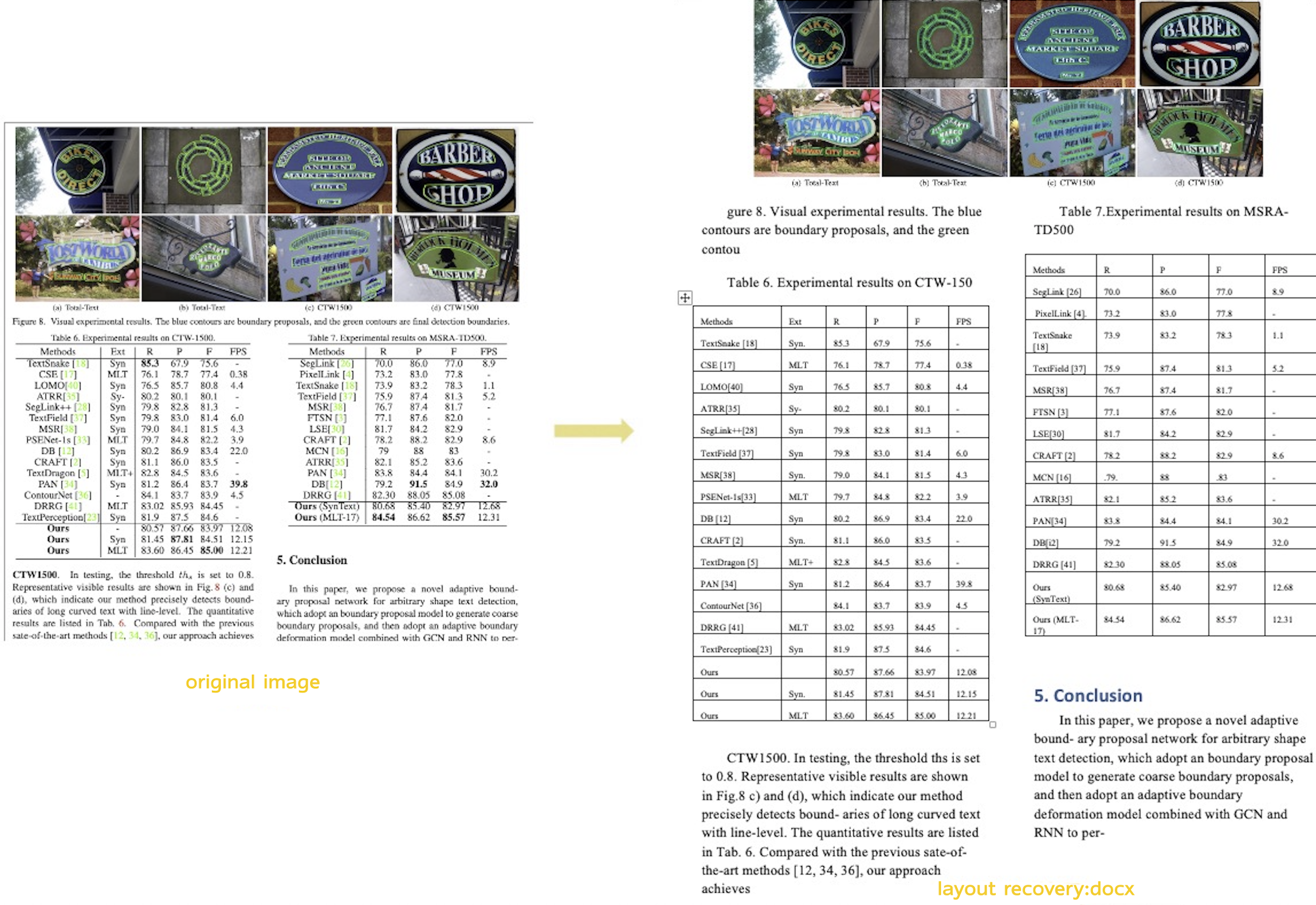}
\caption{Layout Recovery result in PP-StructureV2.}
\label{pic:layout_recovery}
\end{figure}

Layout Recovery a newly added module which is responsible for restoring the image to an editable Word file according to the analysis results. Layout of the restored file is consistent with the original image. Figure \ref{pic:layout_recovery} shows a demo result of Layout Recovery.

\subsection{Key Information Extraction}

\begin{figure*}[h]
\centering
\includegraphics[width=0.9\textwidth]{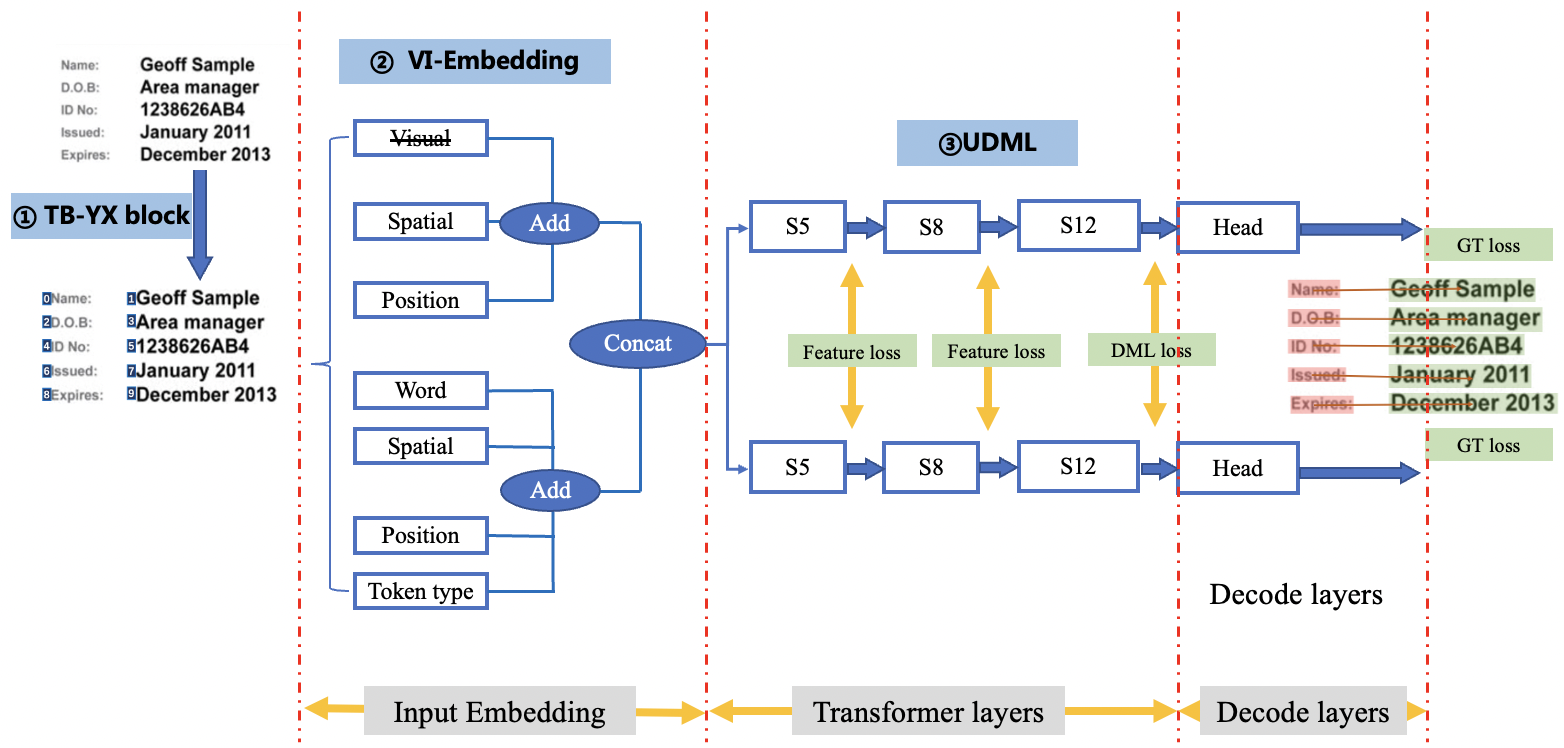}
\caption{Key Information Extraction framework in PP-StructureV2.}
\label{kie_framework}
\end{figure*}

Key Information Extraction (KIE) is usually used to extract the specific information such as name, address and other fields in the ID card or forms. Semantic Entity Recognition (SER) and Relationship Extraction (RE) are two subtasks in KIE, which have been supported in PP-Structure. In PP-StructureV2, we design a visual-feature independent LayoutXLM structure for less inference time cost. TB-YX sorting algorithm and U-DML knowledge distillation are utilized for higher accuracy. Figure \ref{kie_framework} shows the KIE framework.

\subsubsection{VI-LayoutXLM: Visual-feature Independent LayoutXLM}

Visual backbone network is introduced in LayoutLMv2\cite{layoutlmv2} and LayoutXLM\cite{layoutxlm} to extract visual features and combine with subsequent text embedding as multi-modal input embedding. Considering that the visual backbone is base on ResNet\_x101\_64x4d, which takes much time during the visual feature extraction process, we remove this submodule from LayoutXLM. Surprisingly, we found that Hmean of SER and RE tasks based on LayoutXLM is not decreased, and Hmean of SER task based on LayoutLMv2 is just reduced by 2.1\%, while the model size is reduced by about 340MB.

\subsubsection{TB-YX: Threshold-Based YX sorting algorithm}

\begin{figure}[H]
\centering
\subfigure{
\centering
\includegraphics[width=\columnwidth]{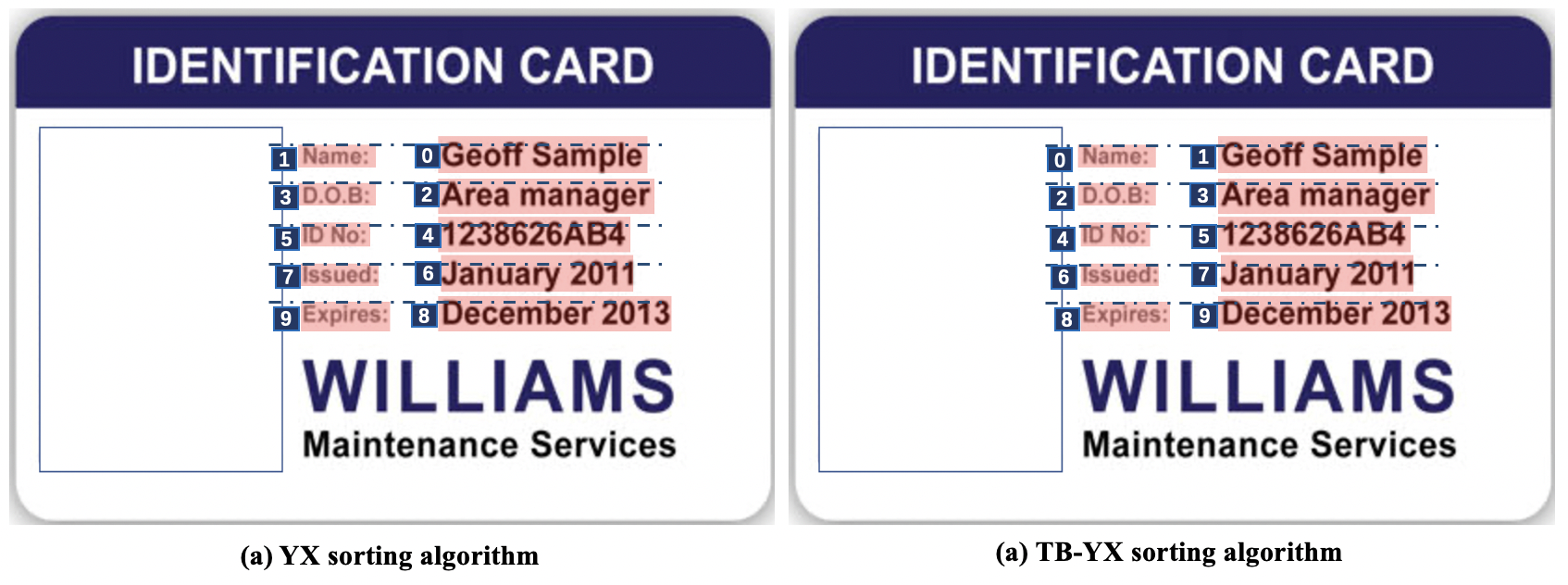}
}
\caption{Results of different sorting algorithms.}
\label{kie_sorting_algo}
\end{figure}

Text reading order is important for KIE tasks. In traditional multi-modal KIE methods, incorrect reading order that may be generated by different OCR engines is not considered, which will directly affect the position embedding and final inference result. Generally, we sort the OCR results from top to bottom and then left to right according to the absolute coordinates of the detected text boxes (YX). The obtained order is usually unstable and not consistent with the reading order as shown in Figure \ref{kie_sorting_algo} (a). We introduce a position offset threshold $th$ to address this problem (TB-YX). The text boxes are still sorted from top to bottom first, but when the distance between the two text boxes in the Y direction is less than the threshold $th$, their order is determined by the order in the X direction, as shown in Figure \ref{kie_sorting_algo} (b). It can be found that the sorted order by TB-YX is more consistent with reading order.

\subsubsection{U-DML: Unified-Deep Mutual Learning}

U-DML is a distillation method proposed in PP-OCRv2\cite{ppocrv2} which can effectively improve the accuracy without increasing model size. In PP-StructureV2, we apply U-DML to the training process of SER and RE tasks, and Hmean is increased by 0.6\% and 5.1\%, repectively.

\section{Experiments}
\subsection{Experimental Setup}
\subsubsection{Datasets}

For Layout Analysis, experiments are carried out on PubLayNet dataset\cite{publaynet}. PubLayNet is a large-scale dataset of document images, which contains 335,703 training, 11,245 validation and 11,405 testing images. Document layout elements such as text, title, list, table and figure are covered. MAP(Mean Average Precision) is used to evaluate the model performance. To verify the strategy generalization, we also carry out experiments on CDLA dataset\cite{cdla}, which is a Chinese layout analysis dataset and covers document elements such ad text, title, figure, figure caption, table, table caption, header, footer, reference, equation. The dataset contains 6,000 annotated images (5,000 for training and 1,000 for validation). 

For Table Recognition, we conduct experiments on PubTabNet\cite{PubTabNet} dataset to verify the effectiveness of the proposed SLANet. PubTabNet contains 500,777 training, 9,115 validation, and 9,138 testing images generated by matching the XML and PDF representations of scientific articles. Since the annotations of the testing set are not released, we only report results on the validation set. A new Tree-Edit-Distance-based Similarity (TEDS) metric for table recognition task is proposed in this work, which can identify both table structure recognition and OCR errors. However, taking OCR errors into account may cause unfair comparison because of different OCR models. Some recent works [\cite{TableStruct-Net}, \cite{Lgpma}, \cite{gte}] have proposed a modified TEDS metric named TEDS-Struct to evaluate table structure recognition accuracy only by ignoring OCR errors. We use accuracy, TEDS and this modified metric to evaluate our approach on this dataset.

For Key Information Extraction, experiments are carried out on XFUND dataset\cite{xfund_dataset}. XFUND\cite{xfund_dataset} is a multilingual form understanding benchmark dataset that includes human-labeled forms with key-value pairs in 7 languages (Chinese, Japanese, Spanish, French, Italian, German, Portuguese). Here, we use Chinese dataset which contains 149 training images and 50 validation images. Hmean is used to evaluate the model performance on both SER and RE tasks. To verify the strategy generalization, we also carry out experiments on FUNSD dataset\cite{funsd_dataset}, which is used for form understanding in noisy scanned documents and contains 199 annotated images (149 for training and 50 for validation).

\subsubsection{Implementation Details}

For Layout Analysis model, we use Momentum with momentum of 0.9 and weight decay 4e-5. Cosine decay learning rate scheduling strategy is adopted with learning rate of 0.4. The batch size and epoch num are set as 24 and 70 on 8*32G V100 GPU devices.

For Table Recognition model, we use Adam optimizer, the initial learning rate is set to 0.001 and adjusted to 0.0001 and 0.00005 after 50 and 60 epochs. The batch size and epoch num are set as 48 and 100 on 4*32G V100 GPU devices.

For Key Information Extraction model, we adopt most of the strategies following \cite{xylayoutlm}. Learning rate, batch size and epoch num are set as $5e^{-5}$, 32 and 200 for SER task, respectively. It's noted that batch size is reduced to 16 for U-DML training process considering the GPU memory. For RE task, the batch size is 8 and the epoch num is set as 130. Constant learning rate strategy with warmup is utilized in RE task for higher accuracy. 4 GPU cards are used for the training process.

\subsection{Layout Analysis}

Ablation experiments on PubLayNet are shown in Table \ref{Ablation_PubLayNet}. PP-YOLOv2 is used for Layout Analysis in PP-Structure. PP-PicoDet-LCNet2.5x is much more efficient than PP-YOLOv2, but mAP is reduced by 1.1\%. By adjusting the input image scale, mAP can be improved by 1.7\%, which is higher than baseline. To get a more lightweight model, we train 1.0x model with FGD, using the previous 2.5x model as the teacher model. The final mAP exceeds the baseline by 0.4\% with the inference speed increasing by 11 times, and the model storage is reduced by 95\%.

To verify the generalization of these strategies, we also conduct ablation experiments on the Chinese Layout Analysis dataset CDLA, and the results are shown in Table \ref{Ablation_CDLA}. It can be found that the performance of layout analysis in both Chinese and English scenarios can be significantly improved.

\begin{table}[t]
\begin{center}
\resizebox{0.45\textwidth}{!}{
\begin{tabular}{l|c|c|c}
\hline
Strategy & \makecell{mAP \\ (\%)} & \makecell{Speed \\ (ms)} &  \makecell{Model \\ Size(M)}\\
\hline
PP-YOLOv2(640*640) &	93.6 &	512 &	221 \\
PP-PicoDet-LCNet2.5x(640*640)  &	92.5 &	53.2 &	29.7 \\
PP-PicoDet-LCNet2.5x(800*608)  &	94.2 &  83.1 & 29.7 \\
PP-PicoDet-LCNet1.0x(800*608)  &	93.5 &  41.2 & 9.7 \\
PP-PicoDet-LCNet1.0x(800*608) + FGD &	94.0 &  41.2 & 9.7 \\
\hline
\end{tabular}}
\end{center}
\caption{Ablation experiments on PubLayNet dataset. \textbf{LCNet} refers to the backbone used in PP-PicoDet. The inference speed is tested on CPU.}
\label{Ablation_PubLayNet}
\end{table}

\begin{table}[t]
\begin{center}
\begin{tabular}{c|c}
\hline
Strategy & \makecell{mAP \\ (\%)}\\
\hline
PP-YOLOv2 &	84.7 \\
PP-PicoDet-LCNet2.5x(800*608)  &	87.8 \\
PP-PicoDet-LCNet1.0x(800*608)  &	84.5 \\
PP-PicoDet-LCNet1.0x(800*608) + FGD &	86.8 \\
\hline
\end{tabular}
\end{center}
\caption{Ablation experiments on CDLA Dataset. LCNet refers to the backbone used in PP-PicoDet. The inference speed is tested on CPU.}
\label{Ablation_CDLA}
\end{table}

We also compare the optimized PP-PicoDet with open source method layout-parser \footnote{https://github.com/Layout-Parser/layout-parser}, which is based on Detectron2. As can be seen from Table \ref{SOTA_PubLayNet}, PP-PicoDet outperforms layout-parser by a large margin on both mAP and inference speed.

\begin{table}[t]
\begin{center}
\begin{tabular}{c|c|c}
\hline
Strategy & \makecell{mAP \\ (\%)} & \makecell{Speed \\ (ms)} \\
\hline

layoutparser(Detectron2) &	88.98 &	2900.0 \\
PP-StructureV2(PP-PicoDet) & 94.00 &  41.2 \\

\hline
\end{tabular}
\end{center}
\caption{Comparison with different methods on PubLayNet dataset.}
\label{SOTA_PubLayNet}
\end{table}

\subsection{Table Recognition}

\begin{table}[t]
\begin{center}
\begin{tabular}{c|c|c|c|c}
\hline
Strategy & \makecell{Acc \\ (\%)} & \makecell{TEDS \\ (\%)} & \makecell{Speed \\ (ms)} &  \makecell{Model \\ Size(M)}\\
\hline

TableRec-RARE &	71.73 &	93.88 &	779 &	6.8 \\
+PP-LCNet  &	74.71 &	94.37 &	778 &	8.7 \\
+CSP-PAN  &	75.68 &	94.72 &	708 &	9.3 \\
+SLAHead  &	77.7 &	94.85 &	766 &	9.2 \\
+MergeToken & 76.31 &	\textbf{95.89} &	766 &	9.2 \\

\hline
\end{tabular}
\end{center}
\caption{Ablation experiments of SLANet on PubTabNet Dataset. The prediction speed is tested on CPU.}
\label{Ablation_SLANet}
\end{table}

\begin{table*}[h]
\begin{center}
\begin{tabular}{l|c|c|c|c|c}
\hline
Methods & \makecell{Acc \\ (\%)} & \makecell{TEDS \\ (\%)} & \makecell{TEDS-Struct \\ (\%)} & \makecell{Inference time \\ (ms)} &  \makecell{Model \\ Size(M)}\\
\hline

EDD & - &	88.3 &	-&	- &	- \\
TableMaster & 77.90 &	96.12 &	- &	2144 &	253 \\
LGPMA  & 65.74 & 94.70 &	96.70 &	- &	177 \\
TableRec-RARE &	71.73 &	93.88 &	- &	779 &	6.8 \\
\hline
SLANet & 76.31 &	95.89 &	\textbf{97.01} &	\textbf{766} &	9.2 \\

\hline
\end{tabular}
\end{center}
\caption{Compare with state-of-the-art methods on PubTabNet dataset.}
\label{SOTA_SLANet}
\end{table*}

Table \ref{Ablation_SLANet} shows the ablation experiments of optimization strategies for SLANet. The baseline model is TableRec-RARE which is proposed in PP-Structure. It can be found that the accuracy can be improved from 71.73\% to 74.71\% by replacing the MobileNetV3 based backbone with PP-LCNet, without increasing the inference time. Using CSP-PAN, the accuracy can be further improved to 75.68\%, and the inference time is reduced by 70ms due to the reduction of the number of feature maps entering the head. Subsequently, we use SLAHead to align the structure and location of cells, which improves the accuracy from 75.68\% to 77.7\%, but the model inference time cost increases from 708ms to 766ms due to the repeated execution of SDM and CLDM. During the previous training processes, the maximum number of tokens can be recognized is set to 500, so images with a token length greater than 500 will not participate in the calculation of the accuracy, but will participate in the calculation of TEDS. After merging tokens that appear in pairs, a HTML string of more tokens can be recognized. Almost all validation sets will participate in the calculation, so the accuracy is reduced slightly, but the TEDS is increased from 94.85\% to 95.89\%. 

We compare our proposed SLANet with several state-of-the-art methods on PubTabNet dataset. Table \ref{SOTA_SLANet} shows the results of SLANet and some state-of-the-art methods on PubTabNet such as EDD\cite{EDD}, TableMaster\cite{TableMaster} and LGPMA\cite{Lgpma}. As can be seen from the table, SLANet is optimal for model size and inference time while maintaining competitive results.

\subsection{Key Information Extraction}

We firstly verify the VI-LayoutXLM's model performance, which can be shown in table \ref{ablation_kie_vi_layoutxlm}. It can be seen that When removing visual feature extraction module in LayoutXLM, the model Hmean is even 0.96\% higher.

\begin{table}[t]
\begin{center}
\begin{tabular}{l|c|c|c}
\hline
Model arch & \makecell{Model size \\ (G)}  & SER Hmean & Gain \\
\hline
LayoutLMv2 & 0.76 & 84.20\% & - \\
VI-LayoutLMv2 & 0.32 & 82.10\% & -2.10\% \\
LayoutXLM & 1.4 & 89.50\% & - \\
VI-LayoutXLM & 1.1 & 90.46\% & \textbf{+0.96\%} \\
\hline
\end{tabular}
\end{center}
\caption{Ablation experiments of on FUNSD dataset.}
\label{ablation_kie_vi_layoutxlm}
\end{table}

The complete ablation experiments are shown in Table \ref{ablation_kie_xfund}. It can seen that reading order of the textlines is vital for the model accuracy, especially for RE task. VI-LayoutXLM is much faster than LayoutXLM under the same condition. Using U-DML knowledge distillation strategy, the model accuracy can be further improved.

\begin{table*}[h]
\begin{center}
\begin{tabular}{l|c|c|c|c|c}
\hline
Strategy & Model size (GB) & SER Hmean & RE Hmean & GPU Inference time (ms) & CPU inference time (ms) \\
\hline
LayoutXLM (baseline) & 1.4 & 89.50\% & 70.81\% & 59.35 & 766.16 \\
VI-LayoutXLM & 1.1 & 90.46\% & 71.87\% & 23.71 & 675.56 \\
+TB-YX & 1.1 & 92.50\% & 78.81\% & 23.71 & 675.56 \\
+U-DML & 1.1 & 93.19\% & 83.92\% & 23.71 & 675.56 \\
+U-DML* & \textbf{1.1} & \textbf{93.19\%} & \textbf{83.92\%} & \textbf{15.49} & \textbf{675.56} \\
\hline
\end{tabular}
\end{center}
\caption{Ablation experiments of on XFUND-zh dataset. Here * means inference using TensorRT.}
\label{ablation_kie_xfund}
\end{table*}

To verify the generalization performance of KIE training strategy, we also carried out experiments on FUNSD dataset \cite{funsd_dataset}, the results are shown in table \ref{ablation_kie_funsd}. More accuracy benefits can be obtained for document images with unordered textlines and noise.

\begin{table}[H]
\begin{center}
\begin{tabular}{l|c|c}
\hline
Strategy & SER Hmean & RE Hmean \\
\hline
PP-Structure KIE & 82.28\% & 53.13\% \\
PP-StructureV2 KIE & \textbf{87.79\%} & \textbf{74.87\%} \\
\hline
\end{tabular}
\end{center}
\caption{Ablation experiments of on FUNSD dataset.}
\label{ablation_kie_funsd}
\end{table}

What's more, we compare our VI-LayoutXLM with several state-of-the-art algorithms on XFUND-zh dataset, which are shown in table \ref{ablation_kie_xfund_sota}. It can be seen that VI-LayoutXLM outperforms most of the multi-modal based methods on XFUND-zh dataset.

\begin{table}[H]
\begin{center}
\begin{tabular}{l|c|c}
\hline
Strategy & SER Hmean & RE Hmean \\
\hline
LayoutLMv2-base KIE & 85.44\% & 67.77\% \\
LayoutXLM-base  & 89.24\% & 70.73\% \\
XYLayoutLM-base & 91.76\% & 74.45\% \\
StrucTexT-large & 92.29\% & \textbf{86.81\%} \\
VI-LayoutXLM-base (ours) & \textbf{93.19\%} & 83.92\% \\
\hline
\end{tabular}
\end{center}
\caption{Comparison with different methods on the XFUND-zh dataset.}
\label{ablation_kie_xfund_sota}
\end{table}

\section{Conclusions}

In this paper, we propose a more robust and comprehensive structural transformation system, PP-StructureV2, which involves 8 improvements. Experiments demonstrate PP-StructureV2 outperforms PP-Structure on all subtasks (Layout Analysis, Table Recognition and Key Information Extraction) in terms of speed and accuracy. The corresponding ablation experiments are also provided.

\bibstyle{aaai21}
\bibliography{eg}

\end{document}